\documentclass{article}

\usepackage{amssymb,dsfont}
\usepackage{amsmath,color,tikz}%
\usepackage{amsfonts}%
\usepackage{graphicx,subfigure}
\usepackage{authblk} 
\providecommand{\U}[1]{\protect\rule{.1in}{.1in}}
\newtheorem{theorem}{Theorem}

\newtheorem{algorithm}[theorem]{Algorithm}

\newtheorem{remark}[theorem]{Remark}

\newcommand{\beq} {\begin{eqnarray*}}
\newcommand{\eeq} {\end{eqnarray*}}

\def \R{\mathbb{R}}

\usepackage{algorithm}
\usepackage[noend]{algpseudocode}

\usepackage{multirow}

\title{A case study : Influence of dimension reduction on regression trees-based algorithms - Predicting aeronautics loads of a derivative aircraft}
\date{}  
\author[1,2,3]{Edouard Fournier\thanks{ \texttt{edouard.fournier@airbus.com } }}\author[2]{  St\'ephane Grihon\thanks{ \texttt{stephane.grihon@airbus.com } }} \author[1,3]{Thierry Klein\thanks{ \texttt{thierry.klein@math.univ-toulouse.fr  or thierry01.klein@enac.fr} } } 
\affil[1]{Institut de math\'ematique, UMR5219; Universit\'e de Toulouse;  CNRS, UPS IMT, F-31062 Toulouse Cedex 9, France} 
\affil[2]{Airbus France 316, Route de Bayonne, Toulouse France} 
\affil[3]{ENAC - Ecole Nationale de l'Aviation Civile, Universit\'e de Toulouse, France}

\begin{document}
\maketitle

\begin{abstract}
In aircraft industry, market needs evolve quickly in a highly competitive context. This requires adapting a given aircraft model in minimum time considering for example an increase of range or the number of passengers (cf A330 NEO family). The computation of loads and stress to resize the airframe is on the critical path of this aircraft variant definition: this is a consuming and costly process, one of the reason being the high dimensionality and the large amount of data. This is why Airbus has invested since a couple of years in Big Data approaches (statistic methods up to machine learning) to improve the speed, the data value extraction and the responsiveness of this process. This paper presents recent advances in this work made in cooperation between Airbus, ENAC and Institut de Math\'{e}matiques de Toulouse in the framework of a proof of value sprint project. It compares the influence of three dimensional reduction techniques (PCA, polynomial fitting, combined) on the extrapolation capabilities of Regression Trees based algorithms for loads prediction. It shows that AdaBoost with Random Forest offers promising results in average in terms of accuracy and computational time to estimate loads on which a PCA is applied only on the outputs.\\

\noindent
\begin{it}Keywords:\end{it} Regression trees, Aeronautics, Dimensional reduction, Extrapolation \\
\noindent
\begin{it} MSC Classification:\end{it} 62J02, 62-07,  63P30
\end{abstract}

\section{Introduction}
In aircraft industry, market needs evolve quickly in a high competitiveness context. This requires adapting a given aircraft model in minimum time considering for example an increase of range or of the number of passengers such as the A330 family in \cite{air}. In our case study, variants concern the maximum take-off weight of a given aircraft model. Depending on the configuration, the computation of loads and stress, as defined in \cite{aiaa88,hje05},  to resize the airframe is on the critical path of this aircraft variant definition: this is a time consuming (approximately a year for a new aircraft variant) and costly process, one of the reason being the high dimensionality and the large amount of data. Big Data approaches such as defined by \cite{gan07} is mandatory to improve the speed, the data value extraction and the responsiveness of the overall process. This study has been realized during a proof of value sprint project within Airbus to demonstrate the usefulness of statistics and machine learning approaches in the Engineering field. In a previous internal project, it has been shown that the family of regression trees \cite{brei84} works well to predict loads for different aircraft missions in an interpolation context. Thus, we can formulate our problem in this way: is it possible to use dimensional reduction and regression trees-based algorithms to predict loads in an extrapolation context (i.e outside the design space of a certain weight variant) to improve the actual process?\\

\subsection{Industrial context}

An airframe structure is a complex system and its design is a complex task involving today many simulation activities generating massive amounts of data. Such is the case of the process of loads and stress computations for an aircraft (that is to say the calculations of the forces and the mechanical strains suffered by the structure) and can be represented as follows:\\

\begin{figure}[h!]
\centering
\includegraphics[width=10cm]{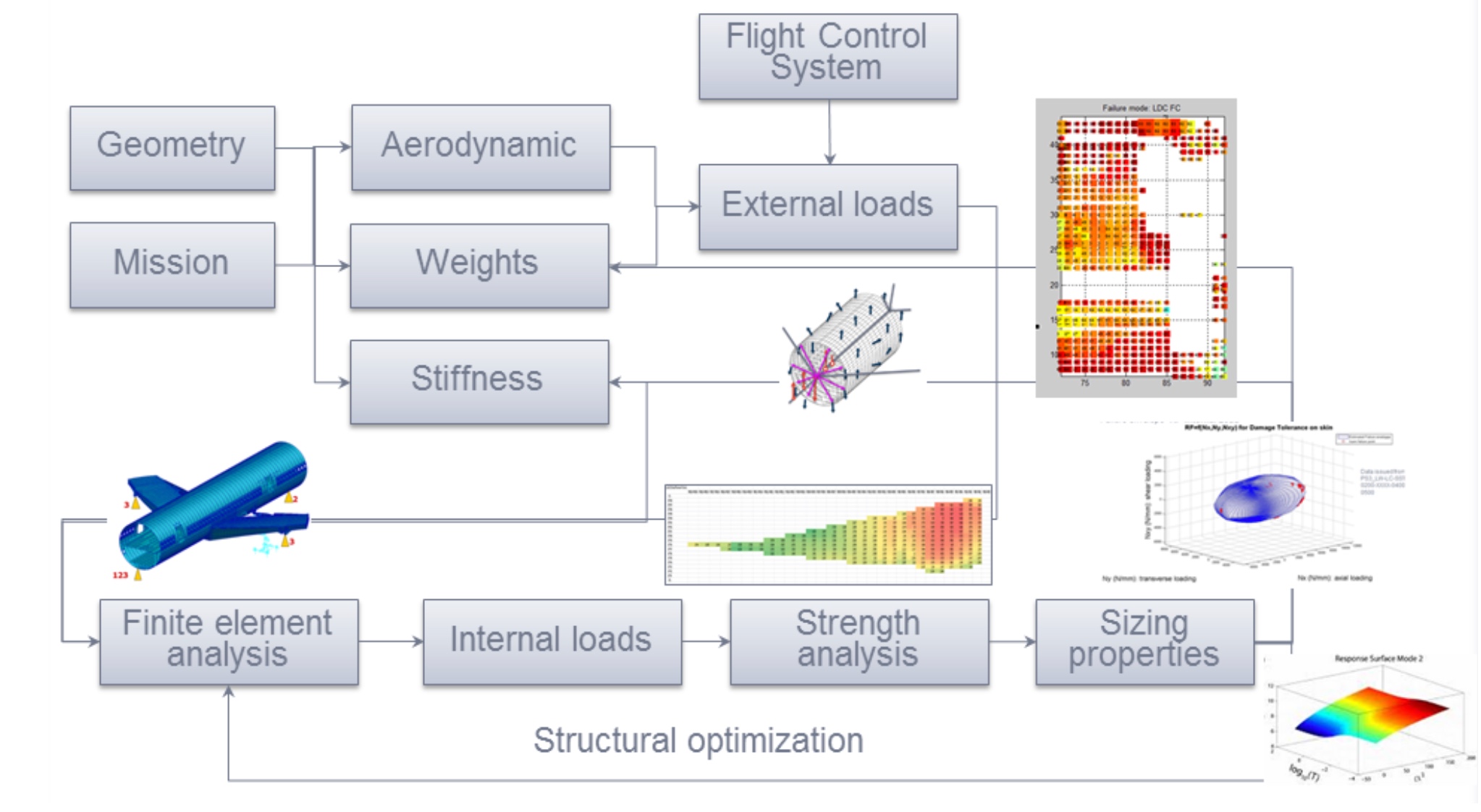}
\caption{Flowchart for loads and stress analysis process}
\label{label-process}
\end{figure}

The overall process exposed in Figure \ref{label-process} is run to identify load cases (i.e aircraft mission and configurations: maneuvers, speed, loading, stiffness...), that are critical in terms of stress endured by the structure and, of course, the parameters which make them critical. The final aim is to size and design the structure (and potentially to reduce loads in order to reduce the weight of the structure). Typically for an overall aircraft structure, millions of load cases can be generated and for each of these load cases millions of structural responses (i.e how structural elements react under such conditions) have to be computed. As a consequence, computational times can be significant.\\

For a derivative aircraft, we can give some rough order of magnitudes in terms of quantities of produced data: External loads ($10^{6}$ of bytes); Weights: number of elements ($10^{4}$ of bytes); Internal loads: number of components by the number of external loads by the number of elements ($10^{11}$ of bytes); Reserve Factors: number of internal loads by the number of failure modes ($10^{12}$ of bytes). Hence, we easily reach $10^{18}$ to $10^{21}$ of bytes for a single derivative aircraft.\\

In an effort to continuously improve methods, tools and ways-of-working, Airbus has invested a lot in digital transformation and the development of infrastructures allowing to treat data (newly or already produced). The objective here is to exploit and adapt Machine Learning and optimization tools in the right places of the computational process.  As pointed by \cite{man11}, these techniques cover a large number of fields such as Internet and Business Intelligence but they can also benefit to the manufacturing industry (here aeronautics). The main industrial challenge for Airbus is to reduce lead time in the computation of loads and preliminary sizing of an airframe.\\

\subsection{A simplistic load and stress model computation process example}

In order to illustrate the process exposed in the previous subsection, let us consider a simplistic load model completed with equations calculating thickness used to correct the weight distribution of a wing structure similar to \cite{doh09}.\\

\begin{figure}[h!]
\centering
\includegraphics[width=10cm]{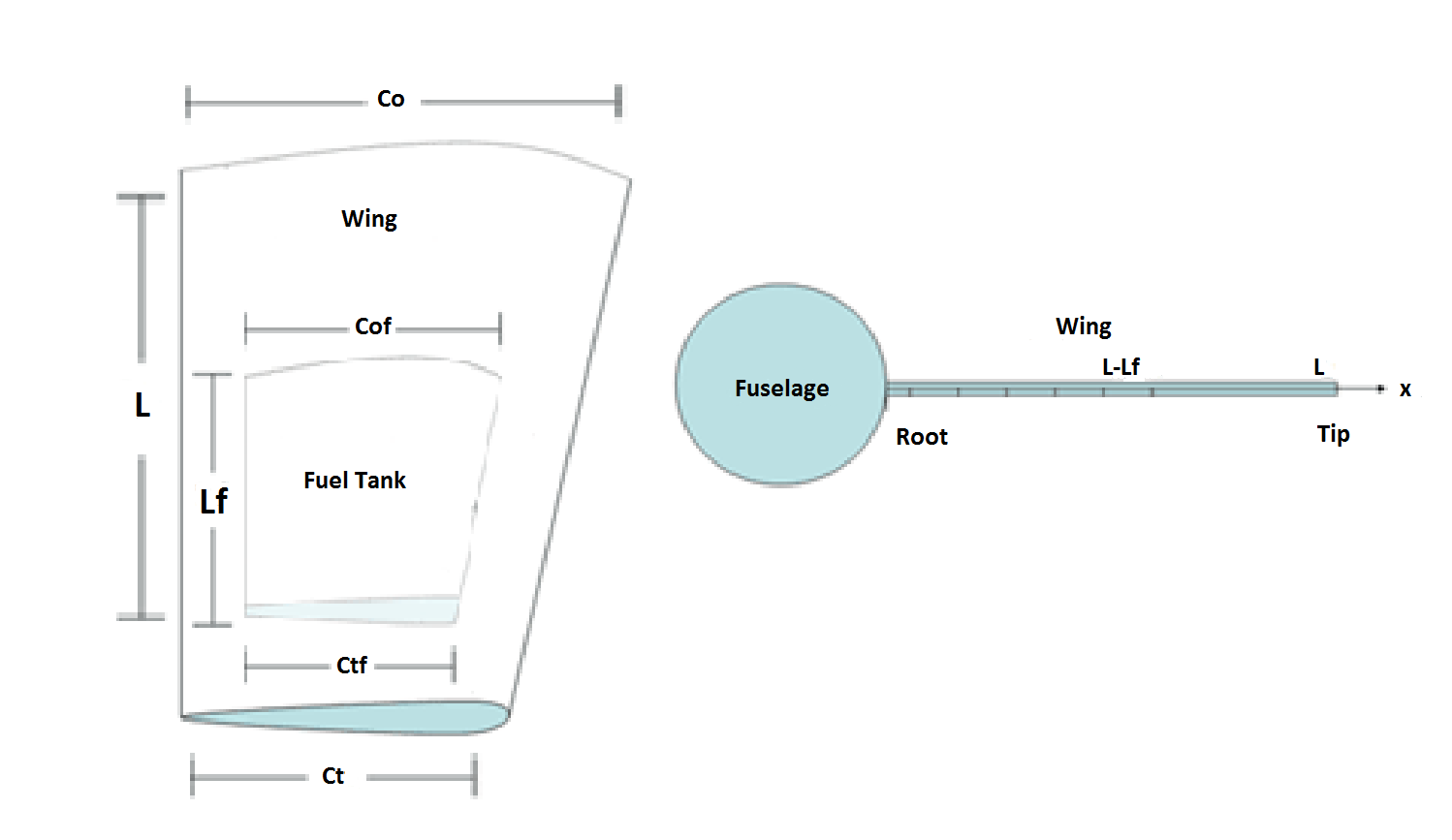}
\caption{Scheme of the wing structure considered in the load model}
\label{label-wing}
\end{figure}

The structure contains a fuel tank at the wing tip with the dimensions Lf, Ctf, Cof as shown in Figure \ref{label-wing}. The length of the wing is L, the chord length at wing root is Co and at the tip Ct. As a consequence, there are three different types of loads which affect the wing: the aerodynamic lift $Q_{lift}$ (i.e the force which allows the aircraft to lift off and to maintain altitude) which depends on the length of the wing, the load factor and the total weight of the aircraft; the loads concerning the fuel and the fuel tank weight $Q_{fuel}$ depending on the fuel weight and the dimension of the fuel tank; and the loads due to the wing structure $Q_{wing structure}$ depending on the weight and the dimension of the wing. By adding these three types of loads, and providing the weight of the wing structure, the weights of the tank and the fuel contained, as well as the total weight of the aircraft and the load factor; $Q_{total}$ provides the basis for calculating the shear force $V$ (transverse forces near to vertical arising from aerodynamic pressure and inertia) and bending moment $M$ (resulting from the shear forces) of the wing. The relations between these quantities are :\\

\begin{center}
$V(x)=-\int_{0}^{L}Q(x)dx$,
\end{center}

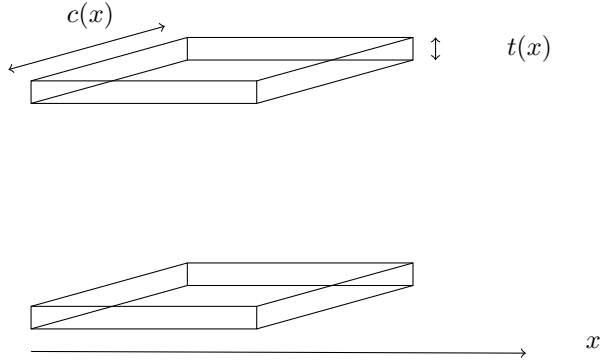
\begin{figure}[h!]
\begin{tikzpicture}[scale=1.5]
%

\draw (1,0,0)--(3,0,0)--(3,0.2,0)--(1,0.2,0)--cycle; 
\draw (0,0,1)--(2,0,1)--(2,0.2,1)--(0,0.2,1)--cycle; 

\draw (1,0,0) -- (0,0,1); 
\draw (3,0,0) -- (2,0,1); 
\draw (3,0.2,0) -- (2,0.2,1); 
\draw (1,0.2,0) -- (0,0.2,1); 

\draw [<-](4,-0.6,0) -- (0,-0.2,1); 
\node[text width=2cm]  at (5.2,-0.5) {$x$};

\draw (1,0+2,0)--(3,0+2,0)--(3,0.2+2,0)--(1,0.2+2,0)--cycle; 
\draw (0,0+2,1)--(2,0+2,1)--(2,0.2+2,1)--(0,0.2+2,1)--cycle; 

\draw (1,0+2,0) -- (0,0+2,1); 
\draw (3,0+2,0) -- (2,0+2,1); 
\draw (3,0.2+2,0) -- (2,0.2+2,1); 
\draw (1,0.2+2,0) -- (0,0.2+2,1); 

\draw [<->](0.8,2.3,0) -- (-0.2,2.3,1);
\node[text width=2cm]  at (0.6,2.4) {$c(x)$};

\draw [<->](3.2,2,0) -- (3.2,2.2,0);
\node[text width=2cm]  at (4.5,2.1) {$t(x)$};
\end{tikzpicture}\\
\caption{Form of the box (upper cover and under cover) of the wing}
\label{label-box}
\end{figure}

\begin{center}
$M(x)=\int_{0}^{L}V(x)dx$,
\end{center}

where $x$ is the position along the wing. We consider that the wing is represented by a simplified rectangular box schematized by two parallel panels representing the covers (see Figure \ref{label-box}) : This is enough to distribute the fluxes induced by the bending moment.\\

We can complete equations calculating thickness. Indeed, by considering the box has height $h(x)$ supposed linearly decreasing along the span, considering we must not exceed an allowable of $\sigma_{max}$ tension and compression. Considering the fluxes in the wing covers are given by $N(x) = \frac{M(x)}{h(x)C(x)}$ thus we have the thickness distribution defined by:\\
\begin{center}
$t(x)=\frac{M(x)}{h(x)C(x)\sigma_{max}}$\\
\end{center}
\newpage
And by integrating we get the weight of the cover given by:\\

\begin{center}
$W_{cover}=2\int_{0}^{L}\frac{M(x)}{h(x)C(x)\sigma_{max}}dx$\\
\end{center}

Indeed, by considering that the wing takes the form of a box presented in Figure \ref{label-box}, by integrating $t(x)C(x)$ along $x$ and by multiplying by $2\rho$, where $\rho$ is the density of the material used to fabricate the wing panels, we get the weight of the wing cover. More precisely, we obtain the minimum weight of the wing cover able to resist an allowable $\sigma_{max}$ tension and compression. We assume that $W_{cover}= 30\% W_{wing}$, then we can extract the minimum weight of the wing structure able to resist an allowable $\sigma_{max}$ tension and compression.\\

\subsection{Data presentation}

The data we have at our disposal are the aircraft parameters (features) which are used in the computing chain for calculating loads (outputs which correspond to moments and forces). We have data coming from the weight variant 238 tons (aircraft parameters and loads distribution along the wing); and we would like to predict those of the 242t and other weight variants (247t and 251t). All the different datasets have been previously computed and we use them to assess the capability of methods defined in the following sections to predict loads in such context. In fact, we hope to answer, by doing so, to the question: "What would the results have been if we had applied such a methodology to calculate the loads instead of the normal process for new weight variants?".\\

25 aircraft (A.C.) parameters play the role of features (lying in $\R$) of a load case and we would like to predict the associated loads (outputs) which are in $\R^{k}$. To simplify, we will focus on predicting bending moment along the wing which is, in our data, represented by a vector of size $k=29$. In other words, each load case (i.e observation) is defined by its 25 features and its bending moment (output). The features are used to identify a typical aircraft event (maneuvers, gusts, continuous turbulences) with specific aerodynamic and weight conditions. Gusts are loads produced by environmental perturbations: sudden vertical or lateral wind blasts which are required by certification organisms like EASA from statistical meteorological histories. Continuous turbulence cases are linked to the cumulative energy stored by the structure under a spectrum of random gusts. A typical maneuver is a 2.5g pull-up consisting in producing an increase aerodynamic lift by deflecting the elevator and increasing the angle of attach of the aircraft. This gives a bending moment close to the maximum value in competition with gust cases. The data base is constituted mainly by gusts (90\% of all load cases) and we will focus on them. To begin, we shall focus on the 238t and 242t data before generalizing our results to other weight variants. A quick summary of the size of our different datasets is presented in Table \ref{table:tab1}:\\
\begin{table}[H] 
\caption{Description of the datasets}
        \begin{tabular}{|c|c|c|c|c|c|c|}
          \hline
         &238t(Train\&Test) & 242t(Validation) \\
          \hline
          \hline
        Dimension data features & 28391 rows x 25 col. & 28391 rows x 25 col. \\
           \hline
        Dimension data outputs & 28391 rows x 29 col. & 28391 rows x 29 col. \\
           \hline
            \hline
        \end{tabular}\\
    \label{table:tab1}
\end{table}
In a more formal way, let be the 238t database of features defined by $\textbf{X}=(X^{1},...,X^{25})$ where $X^{j}$ are quantitative variables (i.e a A.C. parameter), and $X^{j}=(x^{j}_{1},...,x^{j}_{28391})^{T}$. The 238t database of outputs is then defined by $\textbf{Y}=(Y^{1},...,Y^{29})$ and $Y^{j}=(y^{j}_{1},...,y^{j}_{28391})^{T}$. Aircraft parameters $\textbf{X}$ (inputs) we have at our disposal in the training data base 238t are described in Table \ref{table:tab2}:\\

\begin{table}[H]
\caption{Description of the 238t dataset}
\makebox[\textwidth][c]{
\begin{tabular}{|c|c|c|c|c|c|c|}
  \hline
Description & Distribution type & Mean & Std & Min & Max \\
  \hline
  \hline
Defl. Left inboard Elevator & Gaussian & 0.015 & 0.034 & -0.116 & 0.108 \\
  \hline
Stabilizer Setting & Mixture of Gaussian (2 modes) & -0.033 & 0.023 & -0.093 & 0.0033 \\
  \hline
Defl. Spoiler 1 Left Wing & Bi Modal & -0.221 & 0.218 & -0.436 & 0 \\
  \hline
Defl. Spoiler 2 Left Wing & Mixture of Gaussian (2 modes) & -0.266 & 0.262 & -0.755 & 0.230 \\
  \hline
Defl. Spoiler 3 Left Wing & Mixture of Gaussian (2 modes) & -0.266 & 0.262 & -0.755 & 0.230 \\
  \hline
Defl. Spoiler 4 Left Wing & Mixture of Gaussian (2 modes) & -0.266 & 0.262 & -0.755 & 0.230 \\
  \hline
Defl. Spoiler 5 Left Wing & Mixture of Gaussian (2 modes) & -0.266 & 0.262 & -0.755 & 0.230 \\
  \hline
Defl. Spoiler 6 Left Wing & Mixture of Gaussian (2 modes) & -0.266 & 0.262 & -0.755 & 0.230 \\
  \hline
  Defl. all speed inner Aileron & Gaussian & -0.029 & 0.086 & -0.58 & 0.58 \\
  \hline
Defl. Low speed outer Aileron & Quadrimodal & -0.028 & 0.053 & -0.157 & 0 \\
  \hline
Lower part Rudder Deflection & Gaussian & 0 & 0.011 & -0.072 & 0.072 \\
  \hline
Total A.C. Mass & Multimodal & 195738 & 35428 & 135093 & 238000 \\
  \hline
Mach Number & Multimodal & 0.716 & 0.19 & 0.372 & 0.93 \\
  \hline
True Airspeed & Multimodal & 223 & 50 & 126 & 282 \\
  \hline
Altitude & Multimodal & 6270 & 4519 & 0 & 12634 \\
  \hline
x-location of cg in \% amc & Multimodal & 0.297 & 0.114 & 0.140 & 0.42 \\
  \hline
Thrust(calculated) & Multimodal & 131442 & 157160 & 0 & 415495 \\
  \hline
X-Load Factor & Gaussian & -0.020 & 0.107 & -0.3 & 0.261 \\
  \hline
Y-Load Factor & Gaussian & 0 & 0.08 & -0.306 & 0.307 \\
  \hline
Z-Load Factor & Gaussian &  1.024 & 0.43 & -0.701 & 2.643 \\
  \hline
Fuel Tank mass TANK1L & Multimodal & 392 & 1030 & 0 & 4341 \\
  \hline
Fuel Tank mass TANK2L & Multimodal & 13008 & 12721 & 0 & 36295 \\
  \hline
Fuel Tank mass TANK3L & Multimodal & 1883 & 1377 & 0 & 3087 \\
  \hline
Fuel Tank mass TANK1L & Multimodal & 945 & 1029 & 0 & 2592 \\
  \hline
Left inner engine thrust & Multimodal & 65721 & 78579 & 0 & 207747 \\
   \hline 
\hline
\end{tabular}}
\label{table:tab2}
\end{table}

Contrary to the simplistic load calculation example, real simulations needs much more of information: the first ten variables are linked to the orientation of ailerons, spoilers and the rudder which are directional control surfaces (see Figure \ref{label-airplane}); the x-location of gravity center is an indicator concerning the location of the gravity center along the x-axis; the thrust is a calculated variable corresponding to the force which moves the aircraft forward (contrary to the drag force); and the load factors are global indicators which express the "amount of loads" the structure can withstand. All these features are processed by dynamic flight equations considering the flexible body behaviour of the aircraft through finite element models (Lagrange's equations): for further readings, we refer to \cite{fp09}.\\

\begin{figure}[h!]
\includegraphics[width=8cm]{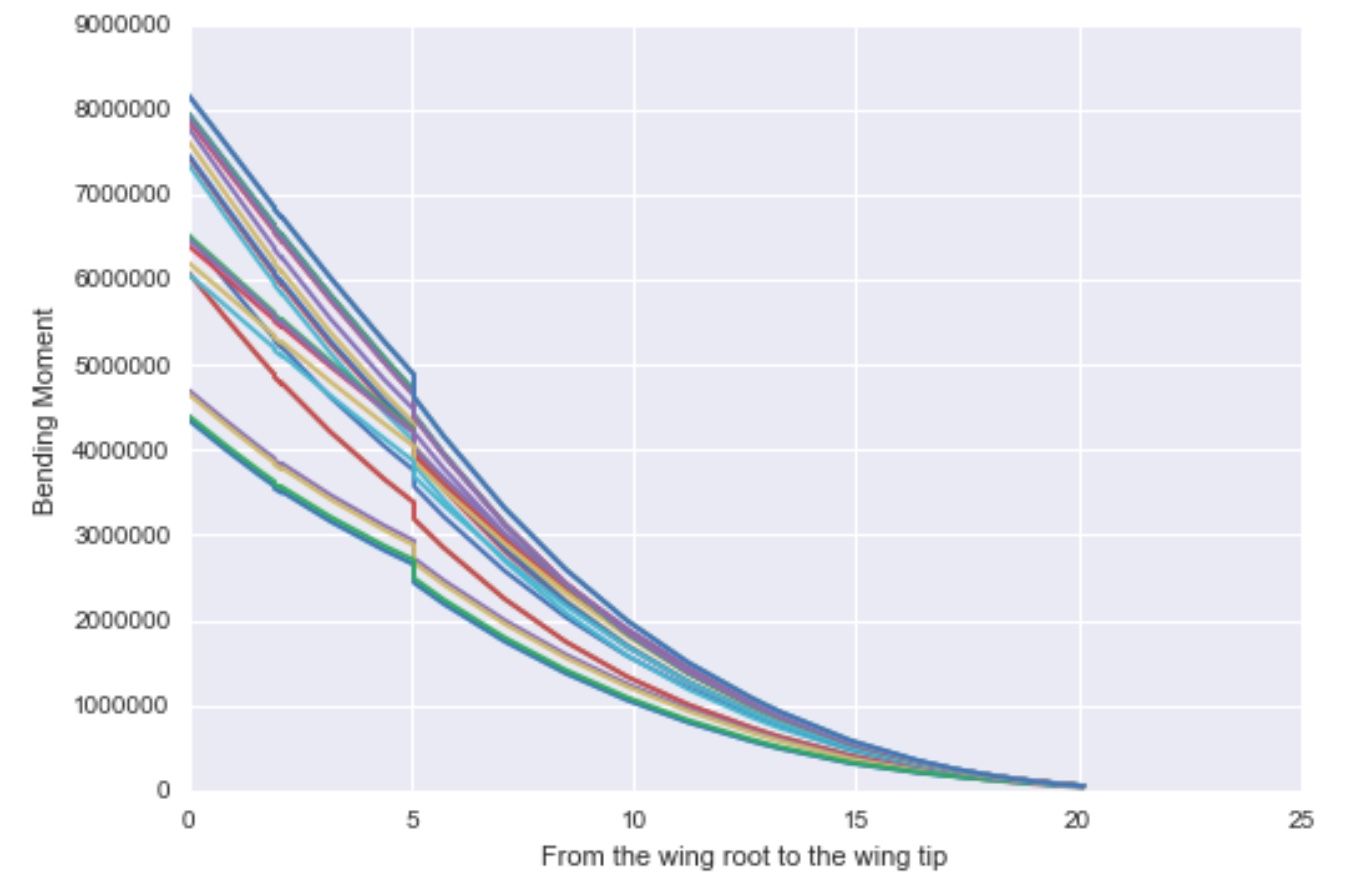}\\
\caption{Examples of bending moments along the wing for different load cases}
\label{label-BM}
\end{figure}

The bending moment is calculated at 29 points along the wing - each point represents a station and stations are not equidistant (two more stations are located in the center wing box; we prefer to focus here on stations of the wing only). Thus $Y^{k}$ represents the values of the bending moment taken at the $k^{th}$ station. Through a change of coordinate system (aircraft system to wing system), we can easily plot bending moments (Figure \ref{label-BM}):

\begin{figure}[h!]
\includegraphics[width=14cm]{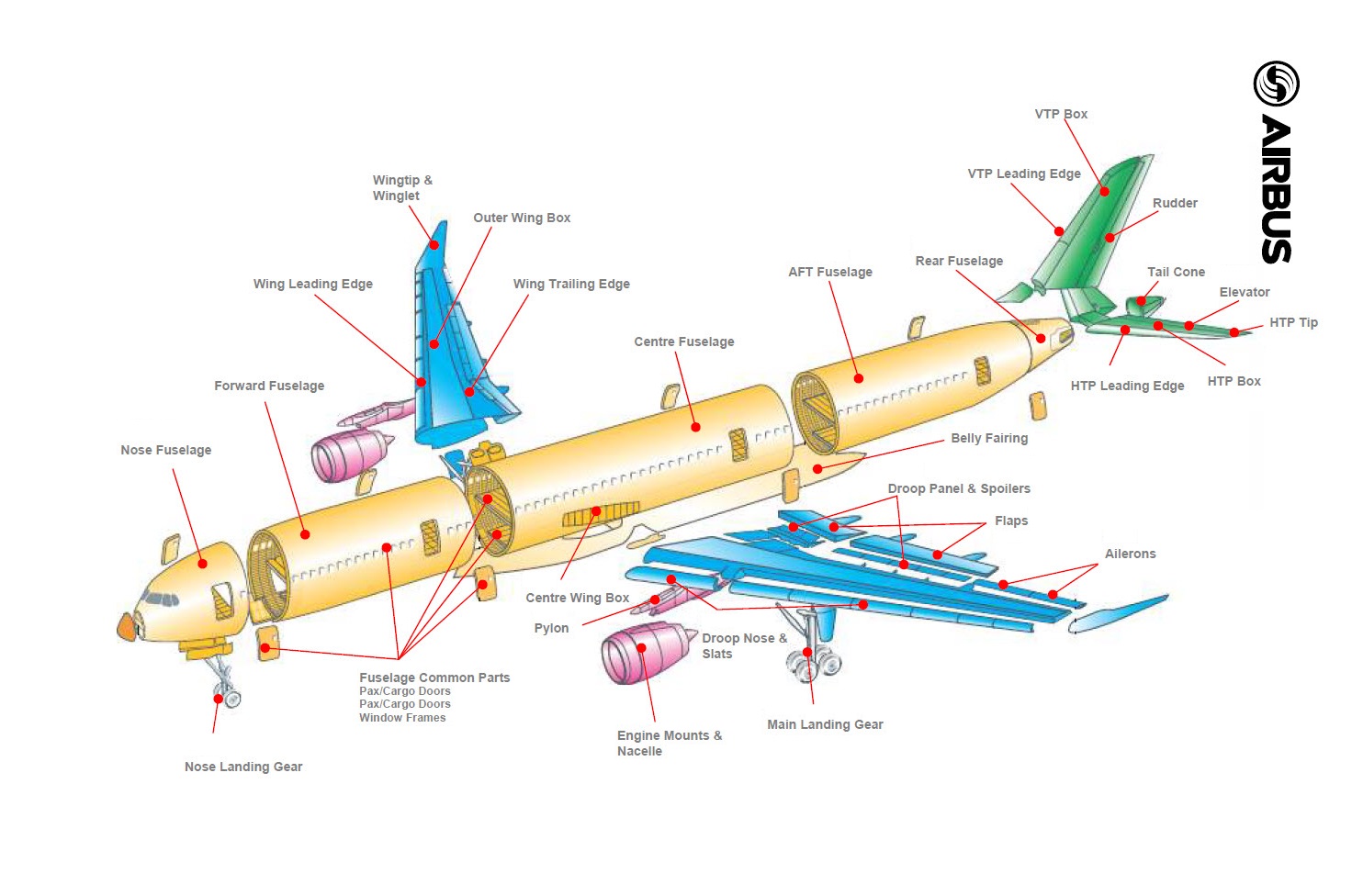}\\
\caption{Airplane parts definition}
\label{label-airplane}
\end{figure}

\newpage

\subsection{Industrial problem}

Aircrafts (A.C.) have been developed for different maximum take-off weight (which is one of the many aircraft parameters used in the computing chain to calculate the loads). Because the computation process exposed above for a new aircraft variant (a new weight variant in our case) can reach easily a year, the use of meta-models, optimization and statistic approaches such defined by \cite{gan07} is mandatory to improve the speed and responsiveness of the overall process.\\ 

From this standpoint, we can expose the following problem: for each combination of A.C. parameters corresponding to a load case, and each load case being categorized into a load condition (family of load cases - gusts or maneuvers), can we give an estimation of the loads for different A.C. parameters for new weight variants (242t, 247t and 251t) knowing the loads of the weight variant 238t?\\

The mathematical problem of this project is an extrapolation problem. Is it possible to "extrapolate" loads of the 242 tons, 247t and 251t knowing loads of the 238t by using machine learning? To be more precise, can we find a function depending on aircraft parameters that allows us to estimate/extrapolate to 242t and other weight variants by learning from those of the 238t? In a previous project concerning loads, it has been shown that the family of regression trees works well on the data we have to deal with. As a consequence, different algorithms based on decision trees will be investigated. Besides, because of the dimension of our outputs, how do dimensional reduction techniques affect the capability of extrapolation of machine learning algorithms based on regression trees?\\

This paper is organized as follows: Section 2 is dedicated to the description of the three different techniques of dimension reduction we used in our study. Then in Section 3 we expose the different algorithms based on regression trees and finally we present in Section 4 our results.

\section{Three Dimensional Reduction Techniques}

In order to improve the efficiency and speed of the modeling process, we compare several dimensional reduction techniques. We start by using a classical PCA on the inputs and also on the outputs. Then we consider a polynomial fitting and finally we mix the two methods. These dimensional reduction techniques will reduce the dimension of the output space. Each technique has been used on the 238t, and these allow us to reverse the technique to come back to the original output space easily.
\subsection{Principal Components Analysis}
In few words, the Principal Components Analysis (PCA), developed by \cite{pea01} and formalized by \cite{hot93}  is a statistical method used to compress a matrix $n$ x $p$ of quantitative variables into a smaller rank matrix. This method uses the variance-covariance matrix (or correlation matrix) to extract important factors (few in general) to represent observations in a smaller subspace. As a consequence, each observation is represented by coordinates into new components linked to these factors (this approach is similar to the SVD decomposition).\\

\begin{figure}[h!]
\includegraphics[width=9cm]{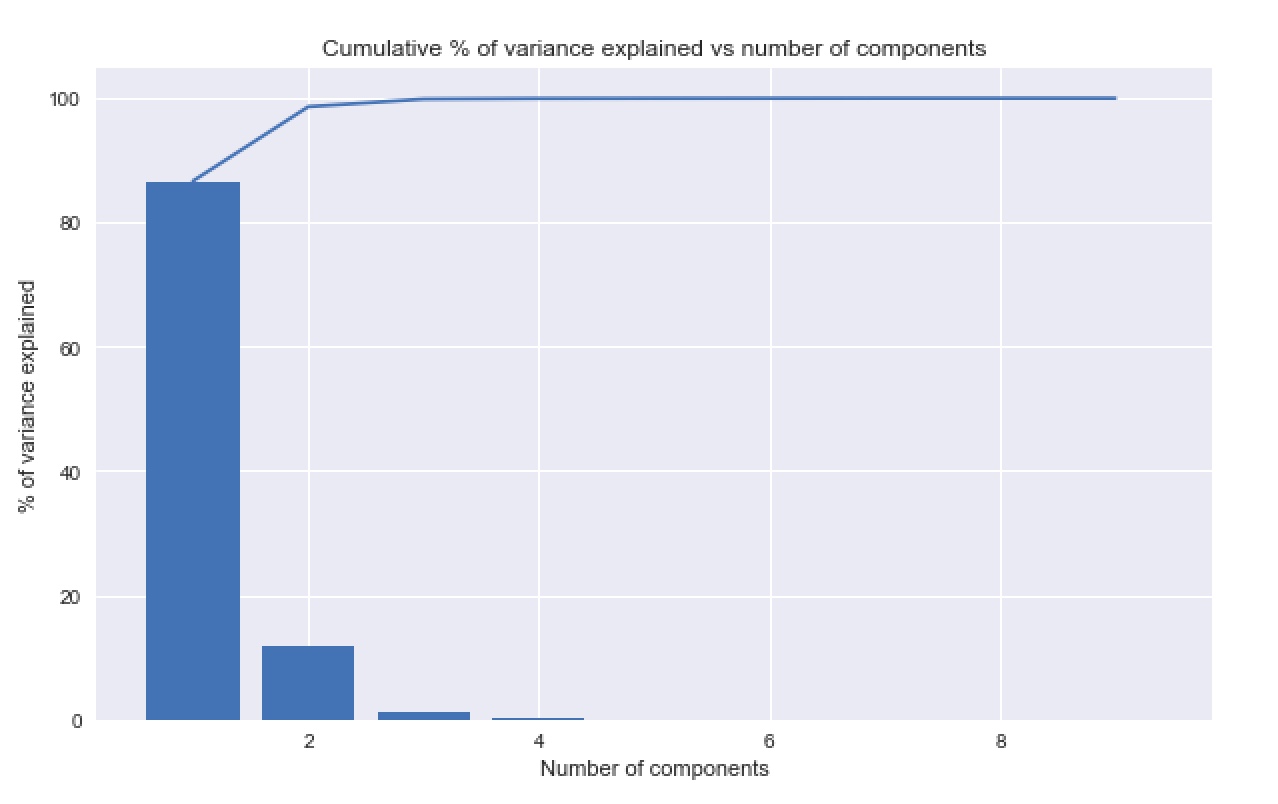}\\
\caption{Cumulative percentage of the explained variance when applying a PCA on the raw outputs}
\label{label-CumVarPCAOut2}
\end{figure}

We apply the PCA in the space defined by the outputs (centered and reduced), and the Figure \ref{label-CumVarPCAOut2} shows the decline of the variance explained by each component as well as the cumulative percentage of the explained variance:\\

The study of the eigenvalues shows that the six first components explain 99.99\% of the total variance. When we look closer at the correlation of the original variables with the principal components, we see that all features have a similar correlation coefficient with the two first principal components.\\

\subsection{Polynomial fitting}

As we can see in Figure \ref{label-BM}, a discontinuity always appears at the 12th station along the wing. Besides, the curves we observe are extremely regular. Consequently, it seems reasonable to fit a polynomial on the first part of the curve and another on the second. In order to choose properly the degree of each polynomial, we assess the quality of the fit by calculating a R-squared score for each curve.

Thus, we consider that it exists a polynomial function $p$ of degree $d$ for each part of the curve such as:
\begin{center}
$p(x)=a_{0}x^{d}+...+a_{d}$\\
\end{center}

The coefficients $a_{0},...,a_{d}$ are obtained by minimizing the squared error by the least squares method.\\

To have an R-squared score greater than 99.9\% for each curve and to avoid over-fitting by choosing too great degrees, the optimal couple of degrees is set to 2 for both polynomials. The dimension of the output space would be 6 instead of 29.

\subsection{Polynomial fitting \& Principal Components Analysis}
By first applying polynomial fitting on the curves and then applying a PCA on the coefficients of the polynomials, we can decrease one more time the dimension of the output space from 6 to 4.\\

By keeping 4 principal components, the output space goes from 6 to the 4 dimensions and the precision is greater than 99.9\% for at least 99\% of the observations. Here follows the decline of the explained variance per component as well as the cumulative percentage of the explained variance (Figure \ref{label-CumVarPolyPCAOut2}):

\begin{figure}[h!]
\includegraphics[width=9cm]{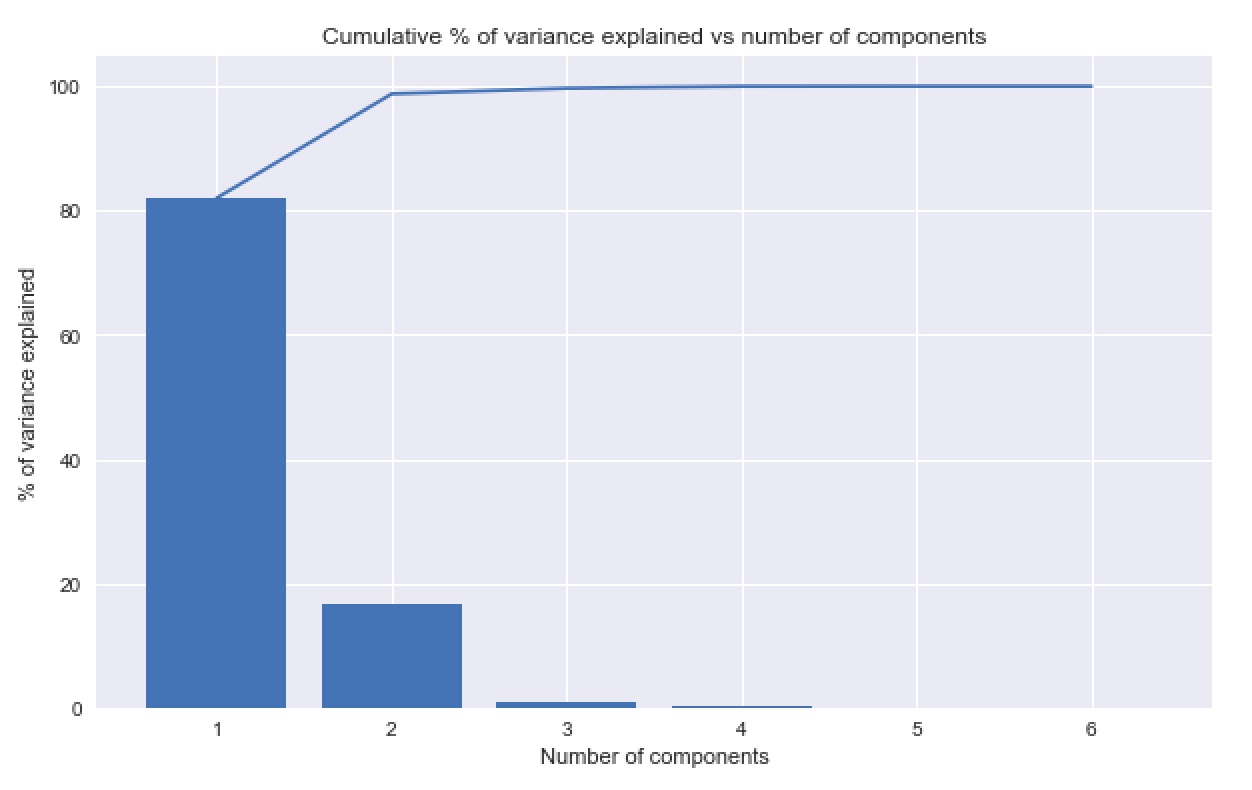}\\
\caption{Cumulative percentage of the explained variance when applying a PCA on the coefficients of polynomials}
\label{label-CumVarPolyPCAOut2}
\end{figure}

In the following, we shall test the different dimensional reduction techniques above which will be compared to no dimensional reduction.\\

\section{Regression based on Trees}

In this section, different algorithms based on decision trees will be investigated. More precisely, the Classification and Regression Trees have been the source of numerous ensemble methods such as Bagging, Random Forest, the Gradient Boosting and AdaBoost and we explain how they work on the data we deal with. Recall we have at our disposal the 238t database of inputs which contains $\textbf{X}=(X^{1},...,X^{25})$ where $X^{j}$ are quantitative variables (i.e a A.C. parameter), and outputs are defined by $\textbf{Y}=(Y^{1},...,Y^{29})$. For each individual, we observe a couple $Z_{i}=(X_{i},Y_{i})$ where $X_{i}=(X^{1}_{i},...,X^{25}_{i})$ and $Y_{i}=(Y^{1}_{i},...,Y^{29}_{i})$. We have thus a sample of observations of size $n=28391$. The aim is to explain $\textbf{Y}$ by a function of $\textbf{X}$. For the sake of simplicity, we will consider the univariate regression $\textbf{Y}^{k}$ (that is to say the value of the bending moment on the $k^{th}$ station) by a function of $\textbf{X}$.\\

\subsection{Classification and Regression Trees (CART)}

Classification and Regression Trees have been formalized by \cite{brei84} and are decision trees. They consist of approximating a function F such as $F: \textbf{X} \to \textbf{Y}^{k}$. This algorithm considers all of 28391 observations and all of the 25 inputs. In no technical terms, the algorithms partitions the data into smaller and smaller sub-samples until all sub-samples are homogeneous in terms of output variables. Let us recall how the method works (see \cite{brei84}, \cite{quin93}):\\

The construction of a tree is the successive partitioning of the output space thanks to the features in the form of a sequence of nodes. At the beginning, the full data set is linked to the initial node (also called the root) and is divided into two classes (two children nodes, left and right) accordingly to a division criteria. Thus, each child node represents a sub-sample of the data-set of the parent node, and recursively from each child node will arise two other children - if a node has no child, it is considered as a terminal node, also called a leaf. The observations belonging to each node must be the most homogeneous, and two children from a node must be the most heterogeneous. In fact at each node, a feature $X^{j}$ is selected and the algorithm finds the threshold of $X^{j}$ (thanks to an impurity measure, also called heterogeneity function or split function) which leads to the most homogeneous sample vs heterogeneous classes. The division criteria leads to know if a node must be a leaf or not, and finally associates each leaf to a value of ${Y}^{k}$.\\

\begin{figure}[h!]
\begin{tikzpicture}[scale=0.3]
\draw (0,0) -- (15,0) ;
\draw (15,0) -- (15,15) ;
\draw (15,15) -- (0,15) ;
\draw (0,15) -- (0,0) ;

\draw (10,0) -- (10,15) ;
\draw (0,10) -- (10,10) ;

\draw (8,10) -- (8,15) ;
\draw [dotted] (8,0) -- (8,10);

\draw (5,5) node[above] {$l_4$};
\draw (12,7) node[above] {$l_3$};
\draw (3,12) node[above] {$l_8$};
\draw (9,12) node[above] {$l_9$};

\draw (10,0) node[below] {$d_1$};
\draw (0,10) node[left] {$d_2$};
\draw (8,0) node[below] {$d_3$};

\draw (7.5,-1.5) node[below] {$X^j$};
\draw (-1.5,7.5) node[left] {$X^k$};

\draw (-12.5,14) circle (1) ;
\draw (-12.5,14) node{$N_1$};
\draw (-12.2,13) -- (-10.5,10.5);
\draw (-12.8,13) -- (-14.5,10.5);

\draw (-9.8,9.8) circle (1) ;
\draw (-9.8,9.8) node{$l_3$};

\draw (-15.2,9.8) circle (1) ;
\draw (-15.2,9.8) node{$N_2$};
\draw (-14.9,8.8) -- (-13,6.3);
\draw (-15.5,8.8) -- (-17.6,6.3);

\draw (-18,5.4) circle (1) ;
\draw (-18,5.4) node{$l_4$};

\draw (-12.6,5.4) circle (1) ;
\draw (-12.6,5.4) node{$N_5$};
\draw (-12.3,4.4) -- (-10.6,1.9);
\draw (-12.9,4.4) -- (-15,1.9);

\draw (-10,1) circle (1) ;
\draw (-10,1) node{$l_9$};
\draw (-15.4,1) circle (1) ;
\draw (-15.4,1) node{$l_8$};

\draw (-9.1,11.8) node{$X^j\geq d_1$};
\draw (-15.7,11.8) node{$X^j\leq d_1$};

\draw (-11.7,7.4) node{$X^k\geq d_2$};
\draw (-19,7.4) node{$X^k\leq d_2$};

\draw (-9.1,3) node{$X^j\geq d_3$};
\draw (-16.2,3) node{$X^j\leq d_3$};

\end{tikzpicture}\\
\caption{Example of construction of a tree \cite{wiksta} : Nodes are designed by $N$, and leaves by $l$}
\label{label-tree}
\end{figure}
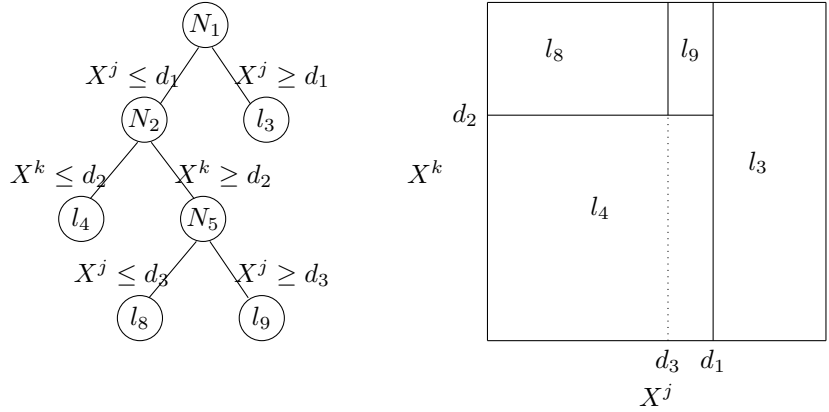

A tree stop growing at a certain node for two reasons: the sub-sample contains too little data according to a fixed threshold set by the user, or the sample linked to the node is homogeneous and no other division is acceptable (that is to say that possible divisions lead to an empty child node). The Figure \ref{label-tree} shows an example of construction of a tree.\\

$N_1$ is the node containing all observations of $\textbf{X}$, and other nodes or leaves contain a subsample of $\textbf{X}$. Let be $\textbf{I}_{l_{j}}:=\{i,\ X_{i}\in l_{j}\}$. Then, the value of $\textbf{Y}^k$ associated to $l_j$ is defined by :\\

\begin{equation}
\textbf{Y}^k_{l_j} = \frac{1}{\# \{\textbf{I}_{l_{j}}\}}\sum_{i\in \textbf{I}_{l_{j}}} Y_{i}^{k}
\label{eq:eq1}
\end{equation}

The value of ${Y}^{k}$ associated to each leaf is then the average value of ${Y}^{k}$s associated to the sub-sample of the leaf.\\

At the end, this algorithm provides a huge tree with many leaves which can lead to over fitting. To avoid this effect, the tree must be pruned: we have to extract a sub-tree. Among a sequence of sub-trees, we keep the one which minimizes a criteria  which depends most of the time of the generalization error and the complexity (the number of leaves): this method is called the cost complexity pruning. In our case, the generalization error (i.e the mean squared error) is calculated by cross-validation.\\

\subsection{Bagging with regression trees}

Bagging is an algorithm which aggregates trees and has been introduced by \cite{brei96}. Let us consider the full sample $\textbf{X}$ of size $n=28391$. For $u=1,...,t$, we denote by $\textbf{X}^{(n,u)}$ a sample of size $n$ obtained by sampling with replacement $\textbf{X}$. For each $\textbf{X}^{(n,u)}$, we train a predictor $p_{u}$. $\{p_{1},...,p_{t}\}$ is therefore an ensemble of predictors, predictors defined on different samples and are tree-based algorithms. Each individual $X_{i},\ i=1,...,28391$, belongs to $t$ differents leaves (one for each tree) denoted by $l_{j_{1}},...,l_{j_{t}}$. So, by equation \ref{eq:eq1}, we have $t$ different values for the prediction of $Y_{i}^{k}$, i.e $(\textbf{Y}_{l_{j_{u}}}^{k})_{u=1,...,t}$. The aggregated prediction value of $Y_{i}^{k}$ is then defined by:
\begin{equation}
\hat{Y}_{i}^{k} = \frac{1}{t}\sum_{u=1}^{t} \textbf{Y}_{l_{j_{u}}}^{k}
\label{eq:eq2}
\end{equation}

Sampling with replacement is most of the time associated to boosting sampling. The method explained above is named Bagging (stands for Boosting AGGregatING). Bagging improves predictions capabilities because it introduces differences between training samples which lead to variability of predictors. Breiman has shown that good candidates to boosting are classification and regression trees and neural networks.

\subsection{Random Forest}

Random Forests, introduced by \cite{brei01}, are based on bootstrap sampling and CART. As in Section 3.2, we first construct $t$ sub samples with replacement of size $n$. When a tree is built, at each node of the tree, we draw randomly $m$ inputs out of 25 (independently) and the optimal splitting criteria is defined through these $m$ drawn variables. Trees grow to the maximal size and are not necessarily pruned.\\

Each tree is an estimator of the underlying function and built on a variation of the training set. As a consequence, each estimator leads to different results. Nevertheless, because of the numbers of estimators, the ensemble of trees (the forest), leads to a stable model. For a new observation, the prediction is then the average value of all the predictions of all predictors as in Bagging.

\subsection{Gradient Boosting}

The gradient boosting, intuited by \cite{brei97} and developed by \cite{frie99}, is like every other boosting method: it combines weak learners. The goal stays the same, to explain $\textbf{Y}^{k}$ by a function of $\textbf{X}$ and instead of tuning parameters of this model, we iteratively add a model to the previous one to increase its capabilities. The name of "gradient" comes from the fact that the gradient of the squared error is the negative residual (see \cite{frie99} and \cite{lii}). In our case, we use regression trees (CART). Here follows a simplified version of the Gradient Boosting Machine algorithm (for more details, see \cite{frie99}):\\

\begin{algorithm}[H]
  \caption{Simplified Gradient Boosting Machine}\label{euclid}
  \begin{algorithmic}[1]
  \Procedure{GBM}{}
  \State Fit a decision tree $F_{1}$ on $\textbf{X}$ (resp. $\textbf{Y}^{k}$) 
  \State Compute the error residuals $e_{1}= \textbf{Y}^{k}-F_{1}(\textbf{X})$
  \For {$t=2,...,T$}:
    \State Fit a decision tree $F$ on $\textbf{X}$ (resp. $e_{t-1}$)
    \State $F_{t}(\textbf{X})=F_{t-1}(\textbf{X})+F(\textbf{X})$
    \State Compute the error residuals $e_{t}= \textbf{Y}^{k}-F_{t}(\textbf{X})$
  \EndFor
  \State The model is then the sum of all fitted trees
  \EndProcedure
  \end{algorithmic}
\end{algorithm}

\subsection{AdaBoost}

One thing that Bagging does not take into account is that each observation is not equally susceptible to be drawn randomly from the training set. Most of the time, we cannot assure this condition. As explained by \cite{druc97}; "in boosting, the probability of a particular example being in the training set of a particular machine depends on the performance of the prior machines on that example". In other words, if machine (a model) is able to predict and learn properly an observation, we do not need to learn more about it, but on observations which are difficult to learn on. Thus, these last ones will be more likely to be picked in a boosting sample. Adaboost was first introduced by \cite{freu95,freu96}, and the following is a slightly modified version by \cite{druc97} called AdaBoost.R2:\\

Initially, each observation is assigned by a weight $w_{i}=1$, $i=1,...,n$. The algorithm is defined this way and continues till the average loss $\overline{L}$ goes under 0.5:
\\

\begin{algorithm}[H]
  \caption{AdaBoost.R2}\label{euclid}
  \begin{algorithmic}[1]
  \Procedure{ADB}{}
  \For {$u=1,...,t$}:
    \State The probability that the observation $i$ is in the training set is directly obtain by $p_{i}=\frac{w_{i}}{\sum w_{i}}$. Draw with replacement a n-sized sample $\textbf{X}^{(n,u)}$ (and its corresponding output $\textbf{Y}^{k}_{u}$) from the training set $\textbf{X}$ (and $\textbf{Y}^{k}$).
    \State Build a model $F_{u}$ on $\textbf{X}^{(n,u)}$ (resp. $\textbf{Y}^{k}_{u}$) by making a weak hypothesis $h_{u}: \textbf{X}^{(n,u)} \to \textbf{Y}^{k}_{u}$
    \State Pass $\textbf{X}$ to the model to get each predictions $F_{u}(X_{i}), i=1,...,n$
    \State Calculate a loss for each observation. The loss may be of any form as long as $L \in [0,1]$
    \State Calculate the average loss:$\overline{L}=\sum_{i=1}^{n}L_{i}p_{i}$
    \State Assessment of the confidence in the predictor by calculating $\beta=\frac{\overline{L}}{1-\overline{L}}$
    \State Update the weights $w_{i} \to w_{i}\beta^{1-L_{i}}$
  \EndFor
  \State Outputs of each machine $F_{u}$ are then weighted, and the predictor is the (weighted) median
  \EndProcedure
  \end{algorithmic}
\end{algorithm}

Although this algorithm is noise and outliers sensitive, it does not need to be calibrated. This ensemble technique can be used with Random Forest and Decision Trees Regressors.\\

\section{Prediction of loads for a new weight variant}
In this section, we apply the techniques we described in Section 3 to our database and present the results we obtain.

\subsection{Data preparation}
Several options are possible to improve the capability of predictions of machine learning. For example, some of them are sensitive to the homogeneousness of the data they learn from, or the number of input variables, as well as outliers. Concerning the last case, we cannot consider outliers because every load cases have been validated thus we must consider all of them. In the first part, we will focus on clustering of our load cases of gusts to improve the ML performance. In the second part, we shall analyze the influence of different dimensional reduction techniques on the generalization capabilities of several algorithms based on regression trees.\\

\begin{figure}[H]
\centering
\includegraphics[width=7.5cm]{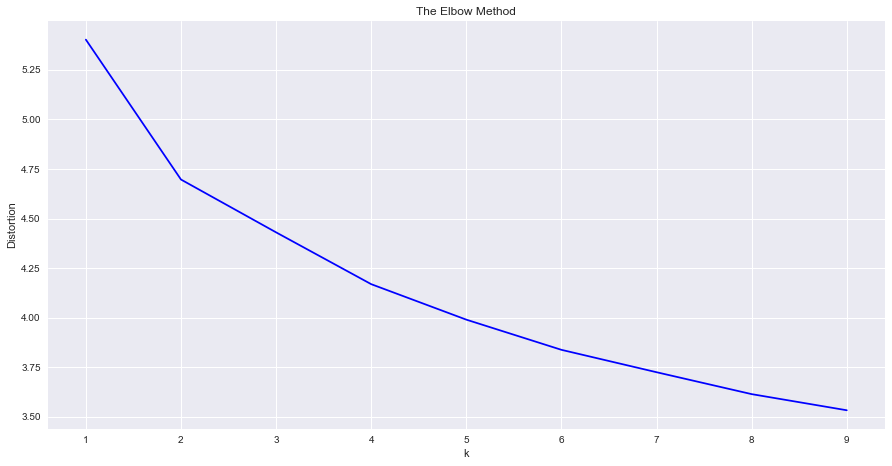}\includegraphics[width=7.5cm]{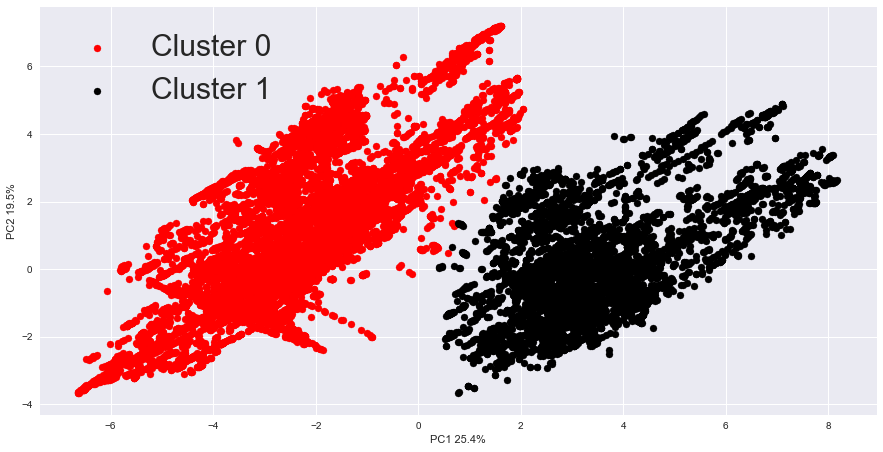}\\

\includegraphics[width=7.5cm]{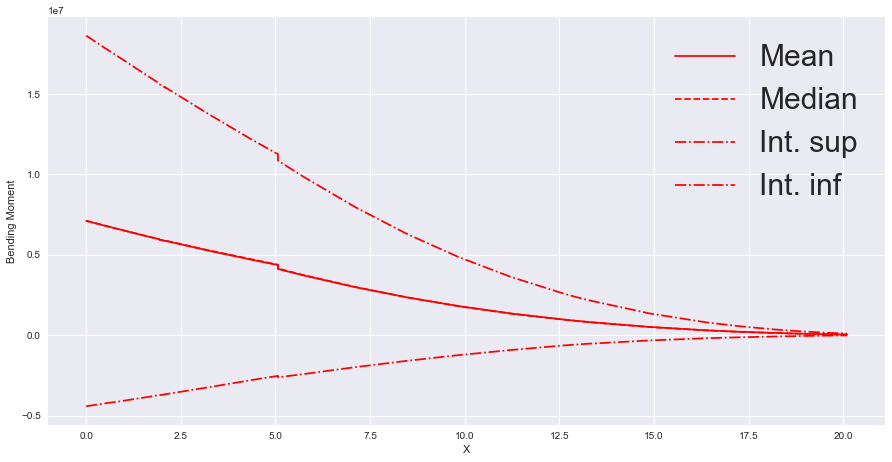}\includegraphics[width=7.5cm]{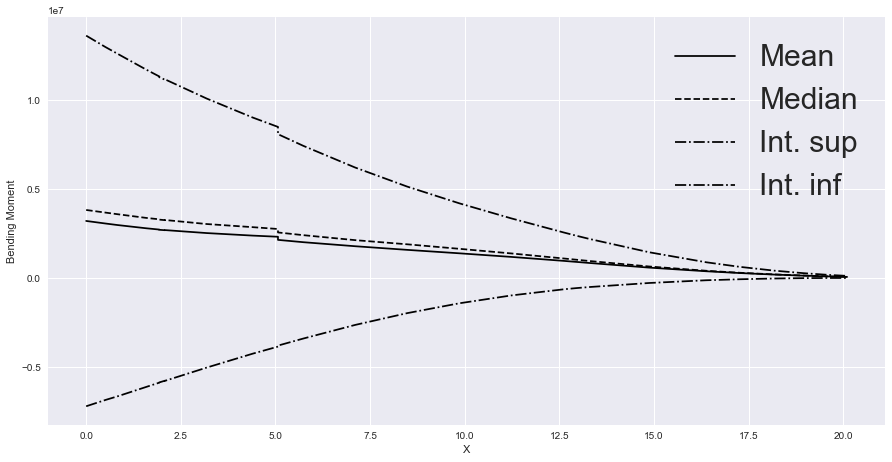}\\

\caption{ (a) Decrease of the Euclidean distorsion according to the number of clusters; (b): Scatter plot of individuals in the two PC; (c)\&(d): Average, median, and Interval Inf. and Sup of  bending moments of Clusters 0 and 1}
\label{label-cluster}
\end{figure}

To improve the capability of machine learning algorithms, clustering has been performed on the gust cases. From a weight variant to another, loads experts are able to roughly estimate the form and intensity of the bending moments. To represent it a priori, we add the coefficients of the polynomials to the features to cluster our data and the K-means algorithm has been performed on these data (features and coefficients). The number of clusters was chosen with the experts and the elbow method using an Euclidean distance. A PCA has been performed and in the two first components, two clusters can be distinguished precisely (see Figure \ref{label-cluster}). In the following, these two clusters will be referred as Cluster 0 and Cluster 1:\\
\newpage

\begin{figure}
\includegraphics[width=10cm]{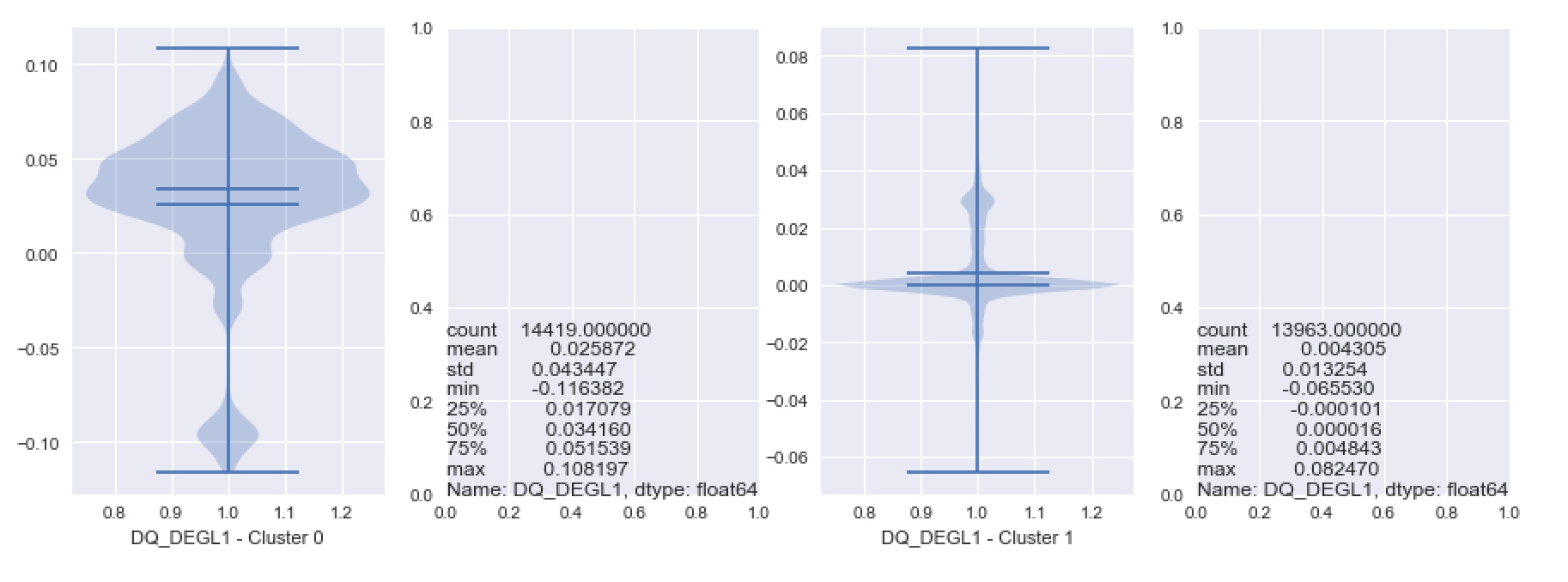}
\caption{Comparison of DQ\_DEGL1(Deflection left inboard Elevator) for the two clusters: the Cluster 0 is mainly constituted by load cases where the left inboard Elevator is active contrary to the Cluster 1}
\label{fig:DQ}
\end{figure}

\begin{figure}[H]
\includegraphics[width=10cm]{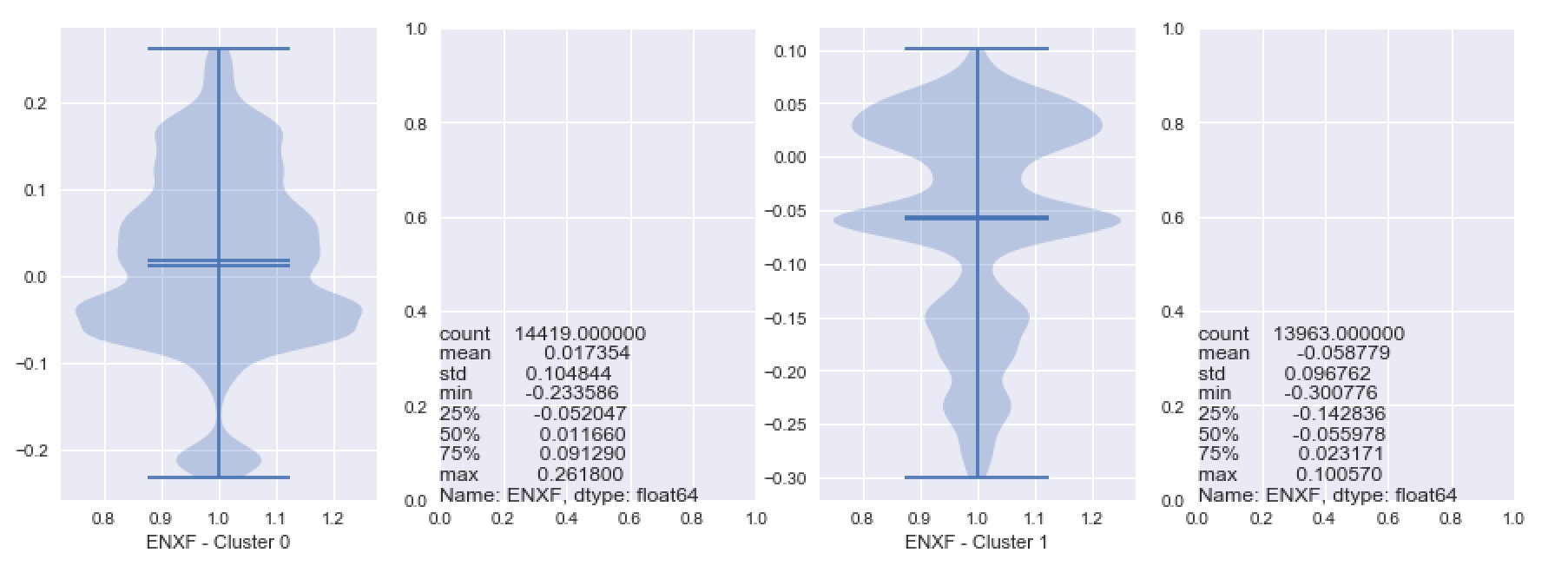}
\caption{Comparison of ENXF (X-Load Factor Body Axis) for the two clusters: the Cluster 0 is mainly constituted by load cases where the X-load Factor Body Axis is positive contrary to the Cluster 1. Simply speaking, that means that the structure "warps" in a way for the Cluster 0, and the other way for the Cluster 1 (due to positive of negative gusts)}
\label{label-NX}
\end{figure}

As we can see in Figure \ref{label-cluster}, the average bending moment of the Cluster 0 is more linear than the one of Cluster 1. Besides, the cluster 1 is constituted by bending moment which are mainly positive and with higher value at the wing root. By looking closer at the A.C. parameters, we can see that most of variables have the same distribution with a slightly different mean value. Nevertheless, some of them are really different (see Table \ref{table:tab3}): this is the case for DQ\_DEGL1 (Deflection left inboard Elevator), DSP\_DEG1L (Deflection Spoiler 1 Left Wing), DP\_DEGIL (Deflection all speed Inner Aileron), DP\_DEGOL (Deflection low speed Outer Aileron) and even more for ENXF (X-Load Factor Body Axis), especially the distribution (see Figure \ref{fig:DQ} and \ref{label-NX}):\\

\begin{table}[H]
\caption{Comparison of variables means in the two clusters: DQ\_DEGL1 (Deflection left inboard Elevator), DSP\_DEG1L (Deflection Spoiler 1 Left Wing), DP\_DEGIL (Deflection all speed Inner Aileron), DP\_DEGOL (Deflection low speed Outer Aileron) ENXF (X-Load Factor Body Axis)}
\begin{tabular}{|c|c|c|c|c|c|c|}
  \hline
 &DQ\_DEGL1 & DSP\_DEG1L & DP\_DEGIL & DP\_DEGOL & ENXF \\
  \hline
  \hline
Cluster 1 & 0.0043 & -0.00025& -0.0082 & -0.0079 & -0.0587 \\
   \hline
Cluster 0 & 0.0258 & -0.4363 & -0.0495 & -0.0488 & 0.0173\\
   \hline
\hline
\end{tabular}
\newline
\newline
\label{table:tab3}
\end{table}
\newpage

\subsection{From 238t to 242t}
Before presenting the results, it is important to explain more the R-squared score we have used in this project and why it is relevant in an engineering context. The R-squared, or also known as coefficient of determination, is a number that shows how well predictions are with respect to the explained variance. In other words, it is a measure of how well the model fits the data:\\

\begin{center}
$R^{2}=1-\frac{\sum_{i}(y_{i}-\hat{y})^{2}}{\sum_{i}(y_{i}-\overline{y})^{2}}$\\
\end{center}

In our case, we calculate a $R^{2}$ at each station of the wing. Indeed, by doing so, we maintain the engineering sense of accuracy of a curve. Because the variance for one curve can be extremely high - for example, we have at the root a value of 8 000 000 and at the wing tip it is closed to 0 - calculating a $R^{2}$ on all the values at the same times would lead to over-estimate the accuracy of our models because the total variance is higher and thus, the ratio between the squared error and the variance is really low.\\

The industrial goal was to have the higher $R^{2}$: in fact, this sprint project is part of a bigger project aiming to deliver models to accelerate pre-development of aircraft. Thus, the necessary condition is to have models precise enough and able to generalize simulations computed anteriorly to approximate, in our case, the computing chain of loads and stress. We agree that the $R^{2}$ can be misleading if the variance of the output is very high. As a consequence, by calculating a $R$-squared at each station (that is to say for each predictor) of the wing: we consider the variance only of the same kind of values in the outputs. The R-squared score given is then the average value of all R-squared calculated at each station.\\

To compare properly the results, from the 238t data set, we have drawn randomly a sample representing 80\% of the observations, the last 20\% represent the test set, and the 242t, 247t and 251t are our validation datasets, and we have repeated the process several time to see if a modification of the training set leads to unstable results in forecasting and generalizing.\\

To perform the comparison of algorithms presented above, we have used the scikit-learn library. Unfortunately, because we are trying to predict a field of vectors (we fit a model per station along the wing), just Random Forest is naturally implemented to do so and to take advantage of links which could exist between them. Simply speaking, when we fit a multioutput model with Random Forest, the impurity measure used at each node has a "covariance" form such as defined in \cite{seg}. Then we used the MultiOutputRegressor for the other algorithms which fits an independent predictor per output vector (i.e per station): the MultiOutputRegressor is then an object containing as much predictors as outputs. As a recall, here are the algorithms we have tested the generalization capabilities: Adaboost based on decision trees regressors (ADB-DT); Adaboost based on Random Forest regressors (ADB-RF), Random Forest (RF), Bagging and Gradient Boosting (GBM). First, before checking the influence of dimensional reduction techniques we check which algorithms work the best on raw data:\\
\begin{table}[H]
\caption{Mean/standard deviation of scores after random learning (80\%) - testing (20\%) - validation: (1) refers to Raw inputs + Raw outputs (no transformation on the data)}
\makebox[\textwidth][c]{
\begin{tabular}{|c|c|c|c||c|c|c||}
  \hline
 &\multicolumn{3}{c|}{Cluster 0} &\multicolumn{3}{c|}{Cluster 1}\\
 \hline
 \hline
 & Learning & Test & Validation 242t &Learning& Test & Validation 242t \\
  \hline
ADB-DT (1) & 0.9999/0 & 0.9756/0.04 &0.956/0.001 & 0.999/0 & 0.983/0.003 & 0.967/0.001 \\
  \hline
ADB-RF (1)& 0.9997/0 & 0.976/0.003 &0.956/0.001 & 0.999/0 & 0.981/0.003 & 0.965/0.001 \\
  \hline
RF (1)& 0.9917/0.003 & 0.96/0.004 &0.92/0.003 & 0.994/0 & 0.966/0.005 & 0.925/0.003 \\
  \hline
Bagging (1)& 0.9922/0.003 & 0.96/0.003 &0.927/0.001 & 0.994/0 & 0.967/0.003 & 0.933/0.001 \\
  \hline
GBM (1)& 0.8858/0 & 0.878/0.004 &0.871/0.007 & 0.896/0 & 0.885/0.003 & 0.878/0 \\
\hline
\hline
\end{tabular}}
\label{table:tab4}
\end{table}

As we can see in Table \ref{table:tab4}, even if AdaBoost is not able to predict and take into account several outputs, the one based on decision tree regressors gets the better results. Random Forest combined with AdaBoost has 3\% higher scores with a lower variability than RandomForest only. It is important to notice that GBM has the less degrowth from the test score to the validation score but the poorest score. Adaboost (based on decision trees or Random Forest) having the best results and the second less degrowth from the test score to the validation score (from 97.56\% to 95.6\%), we will focus on this algorithm to see the impact of dimensional reduction techniques.\\

To quantify the influence of dimensional reduction techniques on extrapolation capabilities, here follows the different configurations we need to compare:\\
\begin{itemize}
\item (1) Raw inputs + raw outputs: no data transformation.\\
\item(2) Raw inputs + PCA outputs: we keep the original input space and we perform a PCA on the
output space.\\
\item(3) Raw inputs + polynomial fitting: we keep the original input space and replace the outputs by
polynomial coefficients.\\
\item(4) Raw inputs + polynomial fitting and PCA: we keep the original input space and replace the
outputs by polynomial coefficients on which we perform a PCA.\\
\item(5) PCA inputs + Raw outputs: we keep the original bending moment and we perform a PCA on
the input space.\\
\item(6) PCA inputs + PCA outputs: we perform a PCA on the design space, and another on the output
space.\\
\item(7) PCA inputs + polynomial fitting: we perform a PCA on the design space and replace the
outputs by polynomial coefficients.\\
\item(8) PCA inputs + polynomial fitting and PCA: we perform a PCA on the design space and replace
the outputs by polynomial coefficients on which we perform a PCA.\\
\end{itemize}

Methods concerning the polynomial fitting are not shown due to lack of generalization and poor results. Other results are shown in Table \ref{table:tab5}.

\begin{table}[H]
\caption{Mean/standard deviation of scores after several random learning (80\%) - testing (20\%) - validation 242t for the configurations: (1) Raw inputs + raw outputs; (2) Raw inputs + PCA outputs; (5) PCA inputs + Raw outputs; (6) PCA inputs + PCA outputs}
\makebox[\textwidth][c]{
\begin{tabular}{|c|c|c|c||c|c|c||}
  \hline
 &\multicolumn{3}{c|}{Cluster 0} &\multicolumn{3}{c|}{Cluster 1}\\
 \hline
 \hline
 & Learning & Test & Validation 242t &Learning& Test & Validation 242t \\
  \hline
(1) ADB-RF  & 0.9997/0 & 0.976/0.003 &0.956/0.001 & 0.999/0 & 0.981/0.003 & 0.965/0.001 \\
  \hline
(2) ADB-RF  & 0.9996/0 & 0.9751/0.002 &0.956/0.0008 & 0.9996/0 & 0.9816/0.003 & 0.966/0.001 \\
  \hline
(5) ADB-RF  & 0.9996/0 & 0.9579/0.004 &0.9120/0.001 & 0.9966/0 & 0.9680/0.004 & 0.9192/0.001 \\
  \hline
(6) ADB-RF  & 0.9995/0 & 0.9585/0.004 &0.9136/0.003 & 0.9995/0 & 0.9684/0.004 & 0.9215/0.002 \\
  \hline
  \hline
(1) ADB-DT  & 0.9999/0 & 0.9756/0.04 &0.956/0.001 & 0.999/0 & 0.983/0.003 & 0.967/0.001 \\
  \hline
(2) ADB-DT  & 0.9998/0 & 0.9742/0.004 &0.9565/0.001 & 0.9998/0 & 0.9823/0.005 & 0.9683/0.001 \\
  \hline
(5) ADB-DT  & 0.9999/0 & 0.9535/0.004 &0.9145/0.001 & 0.9999/0 & 0.9670/0.005 & 0.9141/0.001 \\
  \hline
(6) ADB-DT  & 0.9998/0 & 0.954/0.004 &0.9144/0.003 & 0.9998/0 & 0.9676/0.005 & 0.9247/0.003 \\  
  \hline
  \hline
(1) RF  & 0.9917/0.003 & 0.96/0.004 &0.92/0.003 & 0.994/0 & 0.966/0.005 & 0.925/0.003 \\
  \hline
(2) RF  & 0.9923/0 & 0.9584/0.003 &0.92/0.004 & 0.9937/0 & 0.9658/0.004 & 0.9255/0.001 \\
  \hline
(5) RF  & 0.9889/0 & 0.9407/0.004 &0.8460/0.001 & 0.9899/0 & 0.9475/0.006 & 0.7675/0.001 \\
  \hline
(6) RF  & 0.9889/0 & 0.94/0.004 &0.8681/0.004 & 0.9896/0 & 0.9665/0.004 & 0.7716/0.016 \\  
\hline
\hline
\end{tabular}}
\label{table:tab5}
\end{table}

\begin{remark}
Parameters of algorithms can be consulted in the Appendix A.\\
\end{remark}

PCA performed on the inputs does not improve results but reduces their variability for Random Forest. Nevertheless, we can see that a PCA applied only on the outputs improves slightly the average results when predicting the 242t for all algorithms. This is not surprising that applying a PCA does not highly improve the results since Random Forest and AdaBoost are natively able to deal with a large number of variables.\\

The results of ADB-RF are similar to ADB-DT. One major difference is the variability concerning the validation scores which is reduced against the other methods. From a cluster to another, results concerning the variability and the type of algorithms are the same; just the scores change.\\

AdaBoost with Random Forest or Decision Trees are similar, just the variability in scores is different. Indeed, due to the stable behavior of Random Forests, it is not surprising that AdaBoost performs better on Decision Trees than on Random Forests. Nevertheless, we can assume now that a PCA on the outputs improves the results and from now, we shall investigate how are the error distributed to understand better the lack of generalization capabilities of our model. In the following, just AdaBoost with Random Forest will be investigated concerning the extrapolation with a PCA applied on the outputs.\\

\subsection{From 238t to 251t}

The R-squared is not optimal to appreciate the quality of the fit: this score can hide poor results depending on the data people are dealing with. To assess the goodness of fit of our models, we defined for a curve of bending moment $j$ the error rate as follows:\\

\begin{center}
$error(j)=\sqrt{\frac{\sum_{i=1}^{L}(\hat{y}(x_{i})-y_{j}(x_{i}))^{2}}{\sum_{i=1}^{L}y_{j}^{2}(x_{i})}}$\\
\end{center}

For $j=1,...,n$, where $n$ is the size of the sample we calculate the error rates, and where $L=29$ is the number of stations along the wing. It allows us to have a physical idea of how far our predictions are. For this standpoint, we can easily compute the empirical cumulative distribution function (CDF): $\forall\ j=1,...,n$, let $\alpha \in [0,1]$. The empirical CDF is defined as:\\

\begin{center}
    $\alpha \to G(\alpha)=\frac{1}{n}\sum_{j=1}^{n}\mathds{1}_{(error(j)\leq\alpha)}$
\end{center}

\begin{figure}[h!]
\includegraphics[width=8cm]{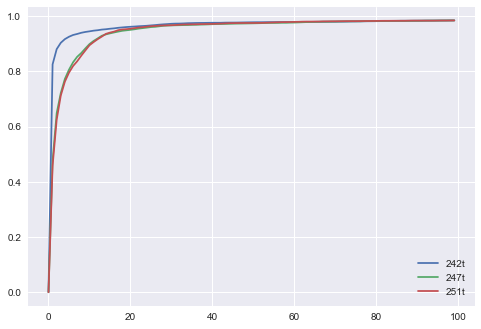}\includegraphics[width=8cm]{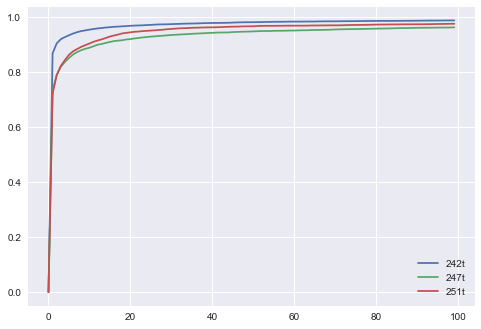}\\
\centering
\caption{Empirical CDF of error rates ($\mathbb{P}(error\leq \alpha)$) concerning the extrapolation for Cluster 0 and Cluster 1: 242t (blue), 247t (green) and 251t (red)}
\label{label-CDF}
\end{figure}

The Table \ref{table:tab6} gives more detailed information concerning the CDF of error rates in Figure \ref{label-CDF}:\\

\begin{table}[H]
\caption{$\mathbb{P}(error\leq 2\%)$, $\mathbb{P}(error\leq 10\%)$ and $\mathbb{E}(error)$ for the different clusters and datasets 242t, 247t and 251t}
\makebox[\textwidth][c]{
\begin{tabular}{|c|c|c|c||c|c|c||}
  \hline
 &\multicolumn{3}{c|}{Cluster 0} &\multicolumn{3}{c|}{Cluster 1}\\
 \hline
 \hline
 & 242t & 247t & 251t &242t & 247t & 251t \\
 \hline
$\mathbb{P}(error\leq 2\%)$  & 88\% & 65\% &63\% &90\% &79\%  &78\%  \\
 \hline
$\mathbb{P}(error\leq 10\%)$  & 95\% &89\%  &89\% &95\%  &88\%  &90\%  \\
 \hline
$\mathbb{E}(error)$  & 12\% & 17\% &22\% & 23\% & 18\% & 27\% \\
 \hline
 \hline
\end{tabular}}
\label{table:tab6}
\end{table}

As soon as we try to generalize our results far from the training dataset, results drop. This is easily explain by the fact that some variables in the 247t and the 251t are far (in average) from the 238t: for example, the quantity of fuel in the first tank is 50\% more important in the 242t, 117\% in the 247t and 270\% more important in the 251t. By looking at Deflection left inboard elevator, it is up to 50\% different in the 247t and 251t than is the 238t and 242t. Unfortunately, theses features have a low importance according to Random Forest (see Appendix B). Besides, it is known that in some cases, slight changes of the features (especially the load factor along the Z-axis) can lead to very different behaviours.\\

\section{Conclusion}

Let us highlight now the contribution of this case study. As mentioned above, AdaBoost associated with Random Forest gives excellent results for observations which are not far from the training set. This is even more accurate when the outputs have similar forms for close design points and for load cases that are not impacted by the weight change roughly. As soon as we try to generalize the results for observations far from the learning data set or for load cases which leads to different behaviour, results drop. If we control the design space at the starting point, or add information concerning the form of the load to predict, or place us in an interpolation context, results would be even better.\\

A PCA on the outputs improves the results in average, and this can be explained because of the high co linearity of the outputs. Because of the presence of outliers and especially because all inputs matter, a PCA on the input space does not improve our results in average.\\

By trying to predict a vector (the shape of our training matrix is 28931x53) and not a point (it would have been 838 999x25), the speed of learning is exponentially decreased, and we keep the engineering information of the mathematical object.\\

Upcoming works concerning this project should investigate the following point: define a reliable method for extrapolation; test other dimensional reduction techniques as the shape invariant model approach such as defined by \cite{ser12} which has been used in the petroleum industry; produce data in sub-spaces where there is a lack of information; investigate the fact that the optimal parameters obtained are maybe not optimal in term of generalization; consider other machine learning algorithms than those based on regression trees because they are known to be not optimal in a generalization problem, because they are considered as ?black-boxes? and because they do not give uncertainties; considering on-line learning: as soon as a new observation is available, the model should keep learning sequentially.\\

Airbus pursues the increasing knowledge capitalization and the development of new methods and tools for Research and Engineering through Big Data initiatives and the promising results of the sprint project, in which this case study has been achieved, are part of the root of upcoming bigger projects about Machine Learning in the load and stress process. 
\newpage
\section*{Appendix A: Models parameters}

Models have been optimize through cross-validation (5-folds). The parameters which do not appear in the following table are set to default value of algorithm in scikit-learn. AdaBoost and Bagging use Decision trees as based estimators: due to time constraints, we have first optimized the parameters of Decision Trees alone on the data, and then optimize AdaBoost and Bagging parameters. Here follows the table containing the parameters of the models exposed in the previous sections.
\begin{table}[H]
\caption{Models parameters through cross-validations (5 folds): models with an asterisk use the parameters of Decision Trees in the same column - (1) Raw inputs + raw outputs; (2) Raw inputs + PCA outputs; (5) PCA inputs + Raw outputs; (6) PCA inputs + PCA outputs.}
\makebox[\textwidth][c]{
\begin{tabular}{|c|c|c|c|c|c||c|c|c|c||}
 \cline{3-10}
 \multicolumn{2}{c|}{}&\multicolumn{4}{c||}{Cluster 0} &\multicolumn{4}{c||}{Cluster 1}\\
 \hline
 \hline
 & Parameters & (1) & (2) & (5) & (6) & (1) & (2) & (5) & (6) \\
 \hline
 \multirow{3}{*}{RF}&min\_samples\_leaf & 5 & 2 & 5 &2 & 3 & 10  & 5 & 3  \\
 \multirow{3}{*}{}&min\_samples\_split & 10 & 11 & 3 & 12 &14  &6  &8 &10  \\
 \multirow{3}{*}{}&n\_estimators & 144 & 192 & 173 & 201 &210  &161  &239 &133   \\
 \hline

 \multirow{5}{*}{ADB-RF}&learning\_rate ADB & 0.95 & 0.92 & 0.98 & 0.96 & 0.90  & 1.07  & 0.92 & 1.06 \\
 \multirow{5}{*}{}&n\_estimators ADB & 31 & 29 & 34 & 25 &34  & 47  & 49 & 38  \\
 \multirow{5}{*}{}&n\_estimators RF & 19 & 23 & 11 & 13 & 22  & 24  & 24 & 12  \\
  \multirow{5}{*}{}&min\_samples\_leaf RF & 17 & 15 & 4 & 3 & 6 & 3  & 2 &9  \\
 \multirow{5}{*}{}&min\_samples\_split RF & 13 & 17 & 4 & 17 & 7  & 7 & 4 &17   \\
 \hline
  \multirow{2}{*}{DT}&min\_samples\_leaf & 4 & 3 & 3 & 2 & 17 & 19  & 15 & 3  \\
 \multirow{2}{*}{}&min\_samples\_split & 9 & 7 & 12 & 7 & 18 & 15 & 12 & 6  \\
 \hline
   \multirow{2}{*}{ADB-DT(*)}&learning\_rate & 1.09 & 0.96 & 1.07 & 0.93 & 1.03 & 0.93  & 0.93 & 1.02  \\
 \multirow{2}{*}{}&n\_estimators & 89 & 117 & 133 & 143 & 156 & 230 & 234  &  172 \\
 \hline
    \multirow{1}{*}{Bagging(*)}&n\_estimators & 186 & 183 & 149 & 146 & 179  & 222   & 216  & 156   \\
 \hline
  \multirow{3}{*}{GBM}&max\_depth & 8 & 10 & 10 &13 &15 &14  &14 &14  \\
 \multirow{3}{*}{}&learning\_rate & 0.92 & 0.90 & 0.90 & 0.97 & 0.91 & 0.99 &0.94 &0.92  \\
 \multirow{3}{*}{}&n\_estimators & 42 & 46 & 52 & 67 & 161 &149  & 69 & 65  \\
 \hline
\end{tabular}}
\label{table:tabA}
\end{table}

One can notice that applying a PCA on the outputs leads to increase significantly the number of estimators in almost all cases when the min\_samples\_leaf and the min\_samples\_split are stable for RF and ADB-RF. Naturally, the number of estimators increases when the depth of the trees grows. A learning\_rate above 1.0 seems to compensate a too large number of estimators and the more transformation we apply on our data, the more deeper are the trees underneath.

\section*{Appendix B: Features importance in Random Forests}

The following table gives the features importance in Random Forest for the cases (1) Raw inputs + raw outputs and (2) Raw inputs +PCA outputs:\\

\begin{table}[H]
\caption{Features Importance in Random Forests - (1) Raw inputs + raw outputs; (2) Raw inputs + PCA outputs}
\makebox[\textwidth][c]{
\begin{tabular}{|c|c|c||c|c||}
 \cline{2-5}
 \multicolumn{1}{c|}{}&\multicolumn{2}{|c||}{Cluster 0} &\multicolumn{2}{c||}{Cluster 1}\\
 \hline
 \hline
  & (1) & (2) & (1) & (2)\\
 \hline
 Defl. Left inboard Elevator & 0.0269  &0.0771  &0.0636  &0.0719\\
 \hline
  Stabilizer Setting & 0.0567  &0.0171  &0.0235  &0.0155\\
 \hline
  Defl. Spoiler 1 Left Wing & 0.0023  & 0.0056 & 0  &0\\
 \hline
  Defl. Spoiler 2 Left Wing &0.0011  &0.001  &0.0001  &0.0001\\
 \hline
  Defl. Spoiler 3 Left Wing &0.0013  &0.0011  &0.0001  &0.0001\\
 \hline
  Defl. Spoiler 4 Left Wing & 0.001 & 0.001 &0.0001  &0.0001\\
 \hline
  Defl. Spoiler 5 Left Wing &0.0012  &0.001  &0.0001  &0.0001\\
 \hline
  Defl. Spoiler 6 Left Wing &0.0012  &0.001  &0.0001  &0.0001\\
 \hline
  Defl. all speed inner Aileron &0.0283  &0.0438  &0.0090  &0.0117\\
 \hline
  Defl. Low speed outer Aileron &0.0114  &0.0158  &0.001  &0.0001\\
 \hline
  Lower part Rudder Deflection &0.0078  &0.0095  &0.0033  &0.0036\\
 \hline
  Total A.C. Mass &0.139  &0.1121  &0.1314  &0.1487\\
 \hline
  Mach Number &0.0052  &0.0145  &0.0074  &0.0064\\
 \hline
  True Airspeed &0.0075  &0.0243  &0.0341  &0.0240\\
 \hline
  Altitude &0.0121  &0.0049  &0.0099  &0.0146\\
 \hline
  x-location of cg in \% amc &0.0039  &0.0086  &0.0082  &0.0067\\
 \hline
  Thrust(calculated) &0.0019  &0.0017  &0.0006  &0.0004\\
 \hline
  X-Load Factor &0.0173  &0.03  &0.0281  &0.0254\\
 \hline
  Y-Load Factor &0.0086  &0.0161  &0.0048  &0.012\\
 \hline
  Z-Load Factor &0.6529  &0.6045  &0.6550  &0.6462\\
 \hline
  Fuel Tank mass TANK1L &0.0016  &0.0013  &0.0028  &0.0023\\
 \hline
  Fuel Tank mass TANK2L &0.0038  &0.003  &0.008  &0.0046\\
 \hline
  Fuel Tank mass TANK3L &0.0015  &0.0012  &0.0042  &0.0024\\
 \hline
  Fuel Tank mass TANK4L &0.0031  &0.002  &0.0033  &0.0014\\
 \hline
  Left inner engine thrust &0.0022  &0.0016  &0.0006  &0.0004\\
 \hline
\end{tabular}}
\label{table:tabB}
\end{table}

Features importance are stable from a method to another and the two most important features are identified: the mass and the Z-load factor. As said in the section 4.3, the importance of variables such as the Deflection left inboard elevator or the quantity of fuel in the first tank is small compared to those last two variables: thus, even if they change roughly for the other weight variants, they have a low impact on the prediction of loads.

\section*{Acknowledgements}
We are very much indebted to the referees and the Associate Editor for their constructive criticisms, comments and remarks that resulted in a major improvement of the original manuscript. We would also like to thank Fabrice Gamboa for careful rereadings.

\bibliographystyle{spmpsci} 
\bibliography{biblio}

\begin{thebibliography}{10}
\providecommand{\url}[1]{{#1}}
\providecommand{\urlprefix}{URL }
\expandafter\ifx\csname urlstyle\endcsname\relax
  \providecommand{\doi}[1]{DOI~\discretionary{}{}{}#1}\else
  \providecommand{\doi}{DOI~\discretionary{}{}{}\begingroup
  \urlstyle{rm}\Url}\fi

\bibitem{air}
Airbus, C.A.: A330 family  (2017).
\newblock Available from :
  http://www.aircraft.airbus.com/aircraftfamilies/passengeraircraft/a330family/

\bibitem{brei96}
Breiman, L.: Bagging predictors.
\newblock Machine Learning \textbf{24}, 123--140 (1996)

\bibitem{brei97}
Breiman, L.: Arcing the edge.
\newblock Technical Report (486) (1997).
\newblock Statistics Department, University of California

\bibitem{brei01}
Breiman, L.: Random forests.
\newblock Machine Learning \textbf{45}, 5--32 (2001)

\bibitem{brei84}
Breiman, L., Friedman, J., Olshen, R., Stone, C.J.: Classification and
  Regression Trees.
\newblock Wadsworth, Belmont, CA (1984)

\bibitem{doh09}
Doherty, D.: Analytical modeling of aircraft wing loads using matlab and
  symbolic math toolbox  (2009)

\bibitem{druc97}
Drucker, H.: Improving regressors using boosting techniques.
\newblock Proceedings of the Fourteenth International Conference on Machine
  Learning pp. 107--115 (1997)

\bibitem{freu95}
Freund, Y., Shapire, R.: A decision-theoretic generalization of on-line
  learning and application to boosting.
\newblock Proceedings of the second European Conference on Computational
  Learning Theory pp. 23--37 (1995)

\bibitem{freu96}
Freund, Y., Shapire, R.: Experiments with a new boosting algorithm, machine
  learning.
\newblock Proceedings of the Thirteenth Conference pp. 148--156 (1996)

\bibitem{frie99}
Friedman, J.H.: Greedy function approximation: A gradient boosting machine
  (1999)

\bibitem{gan07}
Gandomi, A., Haider, M.: Beyond the hype: Big data concepts, methods, and
  analytics.
\newblock Internation Journal of Information Management \textbf{35}, 137--144
  (2015)

\bibitem{hje05}
Hjelmstad, K.D.: Fundamentals of Structural Mechanics.
\newblock Springer US (2005)

\bibitem{aiaa88}
Hoblit, F.M.: Gust Loads on Aircraft: Concepts and Applications.
\newblock AIAA Education Series, AIAA (1988)

\bibitem{hot93}
Hotelling, H.: Analysis of a complex of statistical variables into principal
  components.
\newblock Journal of Educational Psychology \textbf{23}, 417--441 and 498--520
  (1993)

\bibitem{lii}
Li, C.: A gentle introduction to gradient boosting (2016).
\newblock College of Computer and Information Science, Northeastern University.
  Available from:
  http://www.ccs.neu.edu/home/vip/teach/MLcourse/4\_boosting/slides/gradient\_boosting.pdf

\bibitem{man11}
Manyika, J., al.: Big data: the next frontier for innovation, competition and
  productivity  (2011).
\newblock Mc Kinsley Global Institute

\bibitem{pea01}
Pearson, K.: On lines and planes of closest fit to systems of points in space.
\newblock Philosophical Magazine \textbf{2}(11), 559--572 (1901)

\bibitem{quin93}
Quinlan, J.: Programs for machine learning.
\newblock M. Kaufmann (1993)

\bibitem{seg}
Segal, M., Xiao, Y.: Multivariate random forests.
\newblock John Wiley \& Sons, Inc. WIREs Data Mining Knowl Discov \textbf{1},
  80--87 (2011).
\newblock DOI: 10.1002/widm.12

\bibitem{ser12}
Sergienko, E., Gamboa, F., Busby, F.: Shape invariant model approach for
  functional data analysis in uncertainty and sensitivity studies  (2012)

\bibitem{fp09}
Torenbeek, E., Wittenberg, H.: Flight Physics: Essentials of Aeronautical
  Disciplines and Technology, with Historical Notes.
\newblock Springer (2009)

\bibitem{wiksta}
Wikistat: Arbres binaires de d\'{e}cision --- wikistat (2016).
\newblock Available from: http://wikistat.fr/pdf/st-m-app-cart.pdf

\end{thebibliography}
\end{document}